\documentclass{article}

\usepackage[final]{corl_2026} 

\usepackage{url}
\usepackage{xcolor}
\usepackage{subfigure}
\usepackage{graphicx}
\usepackage{amsmath}
\usepackage{wrapfig}

\title{Environment Edits as a Corrective Signal for Reward Learning}

\author{
  Isaac Ngui*\\
  Siebel School of Computing and Data Science\\
  University of Illinois Urbana-Champaign\\
  United States\\
  \texttt{ingui2@illinois.edu} \\
  \And
  Courtney McBeth*\\
  Siebel School of Computing and Data Science\\
  University of Illinois Urbana-Champaign\\
  United States\\
  \texttt{cmcbeth2@illinois.edu} \\
  \And
  Nancy M. Amato\\
  Siebel School of Computing and Data Science\\
  University of Illinois Urbana-Champaign\\
  United States\\
  \texttt{namato@illinois.edu} \\
}

\begin{document}
\maketitle


\begin{abstract}
Robot corrections allow people to adjust a robot's behavior, but a single correction changes many features. Also, preferences about object attributes and relationships between objects cannot be expressed by moving the robot. We propose environment edits as a complementary form of feedback. The robot's motion is held constant as the human edits the scene, so the only explanation for the edit is the moved object. We introduce Feature Learning from Environmental and Robot-Relational Edit Traces (FERRET), which learns preferences from both robot and environment corrections. We show that environment edits are able to disambiguate features and learn preferences that robot corrections alone cannot.
\end{abstract}



\section{Introduction}

Robots that work beside people should support the ability for the people in their surroundings to correct their behavior.
When a robot does something wrong, a human can take the robot arm and move it, indicating how they would like the robot to behave and updating its objective~\cite{bajcsy2017learning, bajcsy2018learning, losey2022physical}.
Corrections like this are quick and they identify preferences about the robot’s motion, such as speed or posture, well.
Feature-learning methods such as Feature Expansive Reward Learning (FERL)~\cite{bobu2021ferl} extend this idea to features that the robot does not already know through asking the human to move the robot along a \emph{feature trace} from where the feature is strongly expressed to where it is not.

A correction is less clear about the world around the robot.
When a person drags a robot arm away from a laptop on a table, the drag also moves the arm from the table edge, lengthens the path, and changes the arm’s posture.
Each of these is a candidate explanation for what the human way trying to convey, and the learner can fit the correction while learning the wrong one.
With several objects on the table, the ambiguity compounds as the correction does not say which object mattered, whether it mattered because of the type of object or state state of the objet, such as a cup being full of water or not.
Some preferences are not able to be expressed by moving the robot in the first place.
“Keep the cup away from the laptop” does not change when the robot moves, so no robot correction carries it. 

We ask what happens if the person corrects the \emph{environment} instead.
The robot’s motion is fixed and replayed while the person edits the scene until the current motion is acceptable.
Interventions can happen on either the robot trajectory or the scene, where a scene intervention leaves every motion feature unchanged, so the edit can only be explained by the object that moved.
These two components are not competitors, but compliments with scene edits capturing preferences about objects, their attributes, and their relations to one another while robot corrections capture preferences about the motion of the robot itself.
Additionally, a robot correction or a scene edit placed at the boundary of acceptability can reveal how much a preference is worth against the cost of moving.

We propose \textbf{FERRET} (\textbf{F}eature Learning from \textbf{E}nvironmental and \textbf{R}obot-\textbf{R}elational \textbf{E}dit \textbf{T}races), which learns preference features from either kind of trace. We explore two different model architectures, one with hand-built relations that capture relationships in the scene, and a graph model that learns important relations within its network.
Our contributions include:
1) an exploration of environment edits as a means of human feedback for preference learning,
2) a proposal of two distinct model architectures for learning human preferences from both robot or environment feature traces,
3) an experimental comparison of these model architectures showing their strengths and weaknesses relative to each other and existing work.


\section{Related Work}
In this section, we discuss prior work related to learning from corrections. We discuss FERL~\cite{bobu2021ferl} in greater detail since it provides the foundation for our work.

\subsection{Learning from Robot Trajectory Corrections}
Corrective demonstrations have long been a tool used to allow humans to express their preference over how a robot should behave in a given environment. Typically these demonstrations are given by physically correcting the robot's behavior. Bajcsy et al.~\cite{bajcsy2017learning, losey2022physical} treat pushes on the robot as evidence of a human's implicit reward function and learn weights online. Losey et al.~\cite{losey2018uncertainty} adds uncertainty over those weights. StROL~\cite{mehta2023strol} improves online learning by using stability analysis to make the learning dynamics more robust to noisy and suboptimal human feedback. Corrections have also been combined with demonstrations and comparative preference rankings in a single learner~\cite{biyik2022learning, mehta2024unified}. All of these methods share the assumption that the human's feedback is the displacement of the robot. We keep that option and add a second one in which the human displaces the scene instead.

\subsection{Ambiguity in Corrections}
A single correction changes many features. Bajcsy et al.~\cite{bajcsy2018learning} showed that updating only the feature that changed most reduced unintended learning. Bobu et al.~\cite{bobu2020misspecification} recognized that scenarios often occur when human corrections may attempt to convey features the robot is unaware of and propose a method for reasoning about how well human feedback is explained by known features. Our approach addresses ambiguity with respect to features related to the environment directly by enabling the user to use the objects to express their preference. Because environment edits hold robot motion fixed, the only explanation for the feature is the object that moved.

\subsection{Feature Expansive Reward Learning}
Feature Expansive Reward Learning (FERL)~\cite{bobu2021ferl} proposes an approach to learning new features previously unknown to the robot using human feedback. Instead of asking for full task demonstrations, the human provides a few \textit{feature traces}, robot trajectories that move from areas of high feature expression to low feature expression. For example, to demonstrate that the robot should not move over a laptop, the human may provide feature traces that start over the laptop and move away. Given a set of feature traces, a neural network learns the feature using a loss based on a Bradley-Terry preference model that encourages feature values to decrease monotonically along each trace and starting and ending feature values for each trace to be indistinguishable. Using feature traces allows FERL to learn more generalizable rewards with small amounts of human data. Our method builds off of FERL and adds support for feature traces consisting of moving objects in the environment.


\section{Feature Learning from Environmental and Robot Relational Edit Traces}
Here, we discuss how we collect feature traces and the different model architectures for FERRET.

\begin{wrapfigure}{r}{0.4\textwidth}
    \begin{center}
    \includegraphics[width=0.8\linewidth]{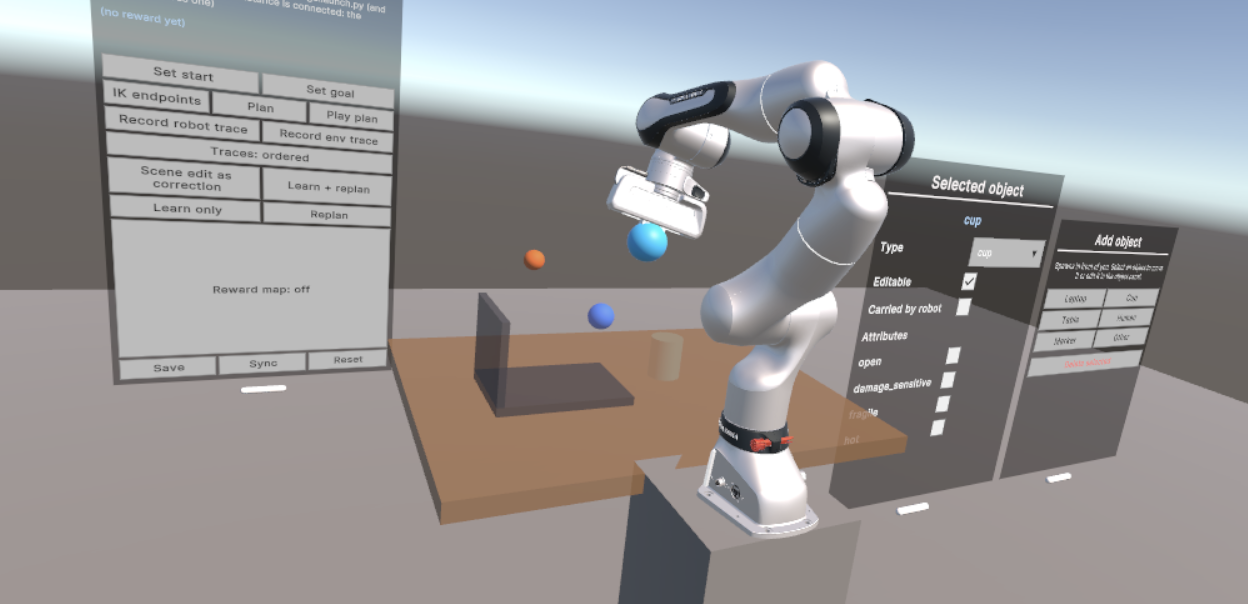}
    \end{center}
    \caption{Our XR system for collecting robot and environment feature traces.}
    \label{fig:vr}
\end{wrapfigure}

\subsection{Collecting Feature Traces in Extended Reality}
Our prior work~\cite{ngui2025rader} explores learning preferences using FERL~\cite{bobu2021ferl} by providing feature traces in extended reality (XR). XR is a beneficial setting to capture feature traces because a virtual robot has no real risk of causing damage while the human places it in high-cost states. When supporting environment traces, XR gives additional benefits. Users can modify characteristics of the objects in the scene (e.g. swap a cup from empty to full) to convey the attributes that are important to the feature they are teaching. We show an example scene in Fig.~\ref{fig:vr}.

\subsection{Learning from Environmental and Robot Edits}
The feature traces used in FERL~\cite{bobu2021ferl} consist of the robot state and the x, y, z positions of objects in the scene. We augment this with additional inputs and explore two different models for reward learning. One, our \textit{relational model}, is a set encoder that operates on end effector-object relation vectors. The other, our \textit{graph model}, is a graph neural network, which uses a scene graph, that moves relational computation into the network rather than relying on hand-designed model inputs.

\subsubsection{Relational Model}
The relational model represents each state as a set of end effector-object relations rather than a single state vector as in FERL~\cite{bobu2021ferl}. For every object, $o \in \mathcal{O}$, in the scene we compute a relation vector $r_o(x)$ consisting of the end effector's position relative to the object, its distance to the object's center, its signed distance to the object's footprint, the object's attribute bits and type, and the attributes of any object being carried by the robot.
Attribute bits indicate, e.g., if the object is open, fragile, or hot. The object type is given by a one-hot vector with options like laptop, cup, and vase. We manually specify these values; however, this could be extended to use object embeddings~\cite{khandelwal2022clip}.

A shared three-layer MLP, $g_{\theta}$, scores each relation independently. The scores are pooled over objects with a softmax and a softplus followed by min/max scaling, yielding the feature:
$\phi_\theta(x) = \operatorname{softplus}\Big( \tau \log \sum_{o \in \mathcal{O}} \exp\big( g_\theta(r_o(x)) / \tau \big) \Big) \in [0, 1]
$
where $\tau$ is the temperature of the softmax; we use a value of 0.5 in our experiments. The feature is permutation-invariant and agnostic to the number of objects. Because we compute relative positions between the objects in the scene and the robot's end effector, displacing an object is equivalent to moving the end effector in the opposite direction, enabling us to learn from correcting either the robot or the scene.

\subsubsection{Graph Model}
The graph model removes the hand-built relation vector and computes relations inside the network. Each state is a scene graph with one node for the end effector, one node per object center containing the object's attribute bits and type, and eight corner nodes per object at the vertices of its footprint. Node embeddings are initialized from the type and attributes. Object position is incorporated through edge features, which consist of the relative position between two nodes and a log-scaled distance, which, like the relational model, makes the feature permutation-invariant and agnostic to the number of objects. Edges run from each corner node to its object center, between object centers, and from every node to the end effector. Each message passing round computes a message from the two endpoint embeddings and the edge attributes, aggregrates incoming messages with attention weights normalized per destination node, and applies a residual update. The end effector embedding is then input to an MLP and passed through the same softplus and min/max scaling as the relational model. Unlike the relational model which only sees end effector-object pairs, the object-object edges let the network represent configurations such as an open cup sitting next to a laptop.

We consider two variants of the graph model which differ in how the relative position on each edge is scaled and in the number of message passing rounds. The \textit{Graph} model scales each relative position by a fixed value and runs two rounds. The \textit{SoftGraph} model divides each relative position by a fixed amount plus the distance itself, so the displacement is approximately preserved near the object but is bounded by one and reduces to a unit vector when far away. SoftGraph runs one message passing round. The Graph model keeps the full amplitude of the displacement, which the network needs to train reliably in a compact workspace, but it lets larger or distant objects dominate the edge inputs. The second round of message passing lets object interact before the end effector reads them. Alternatively, SoftGraph's soft scaling bounds every edge input by one, so no object can dominate by size or distance, making it less prone to taking a large object such as the table as a proxy for absolute position, at the cost of a weaker signal. The single round of message passing limits it to pairwise reasoning. Relative to the relational model, both graph model variants learn the feature from raw geometry with no hand-computed relation, and the object-object edges can express interactions between objects that a pairwise model cannot. However, the ability to mix any object's geometry into the feature can make them less robust than the relational model's hand-built relations.

\begin{figure}
    \centering
    \subfigure[Ground Truth]{
        \includegraphics[height=0.18\textwidth]{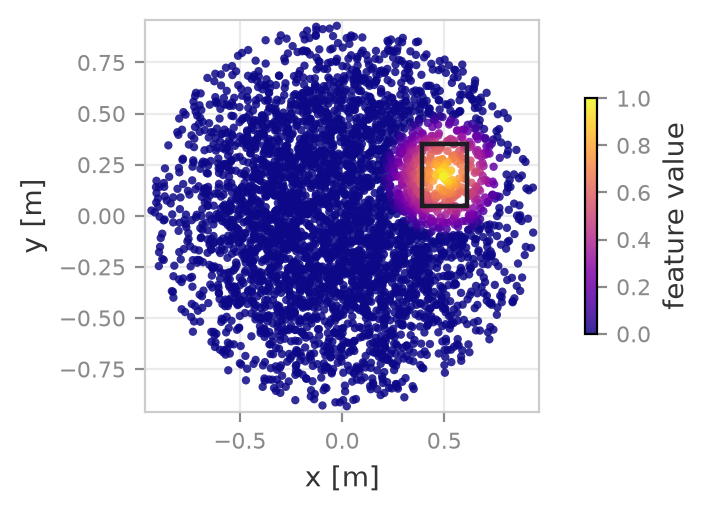}
    }
    \subfigure[FERL Robot]{
        \includegraphics[height=0.18\textwidth]{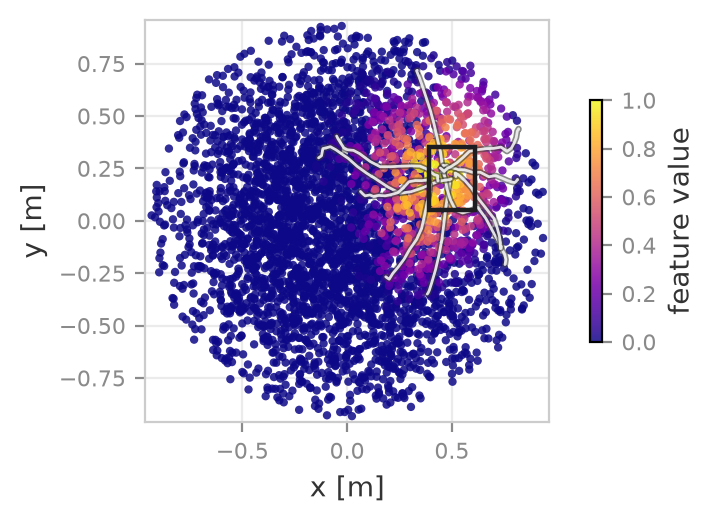}
    }
    \subfigure[FERL Environment]{
        \includegraphics[height=0.18\textwidth]{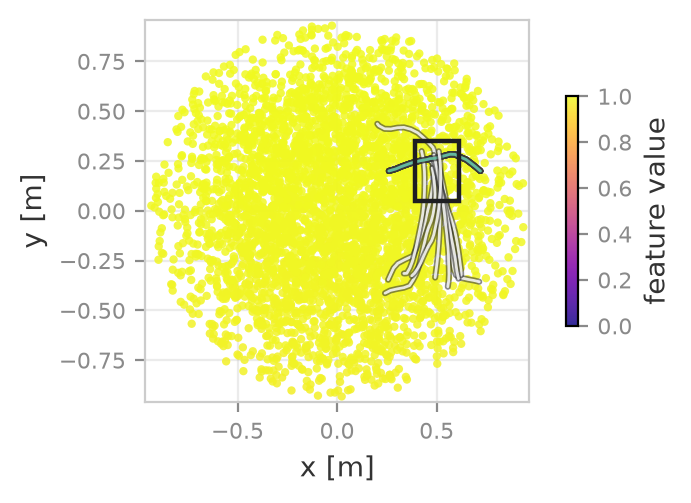}
    }
    \caption{Ground truth and FERL learned features for 'avoid laptop'. Higher feature values indicate lower reward/higher cost. The laptop outline is shown in black. Feature traces, either robot or environment, are shown in white. For environment traces, the robot path is shown in teal.}
    \label{fig:laptop}
    \vspace{-12pt}
\end{figure}

\begin{figure}
    \centering
    \subfigure[Robot MSE]{
        \includegraphics[width=0.3\textwidth]{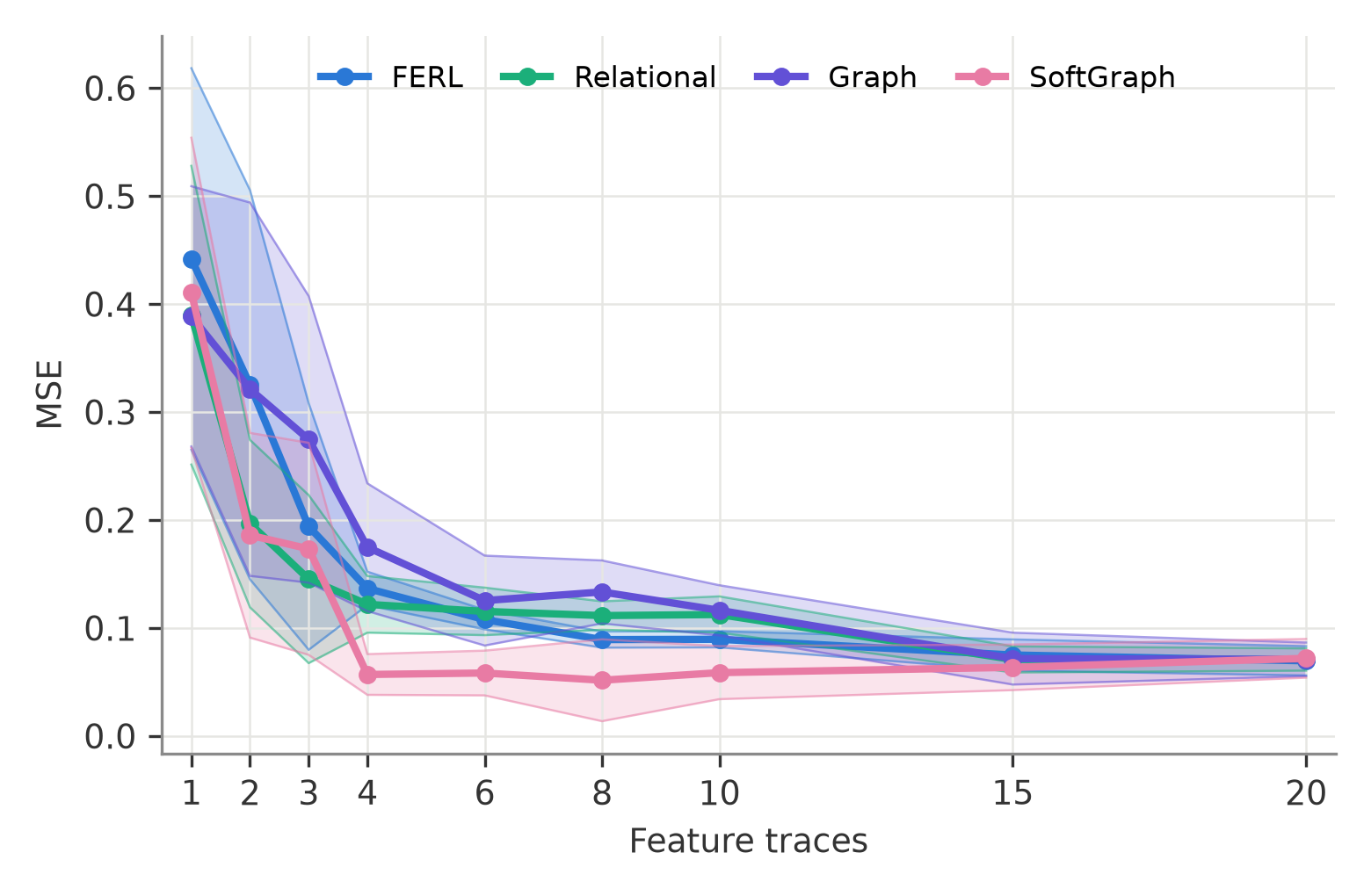}
    }
    \subfigure[Robot PR-AUC]{
        \includegraphics[width=0.3\textwidth]{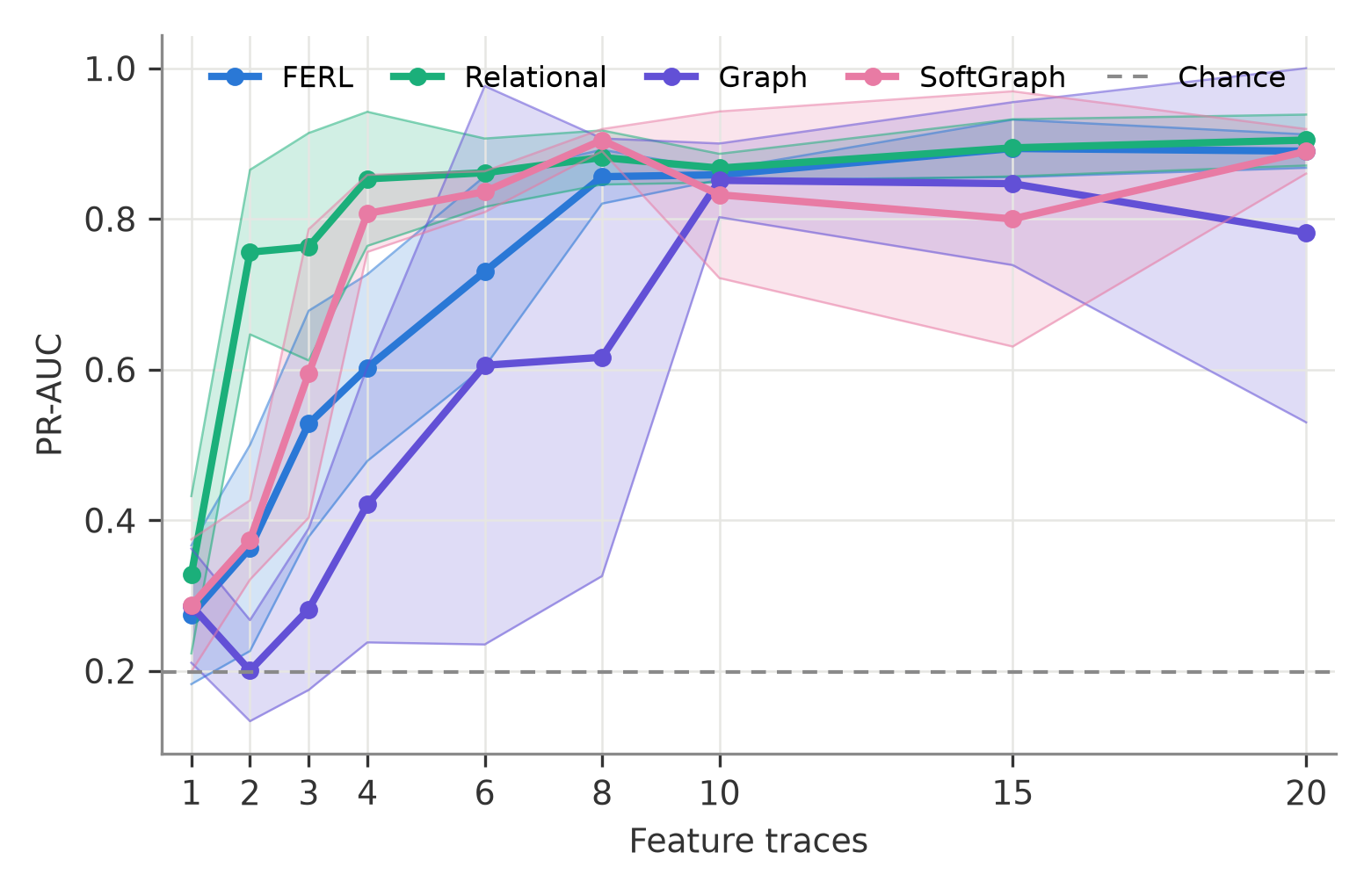}
    }
    \subfigure[Robot Spearman]{
        \includegraphics[width=0.3\textwidth]{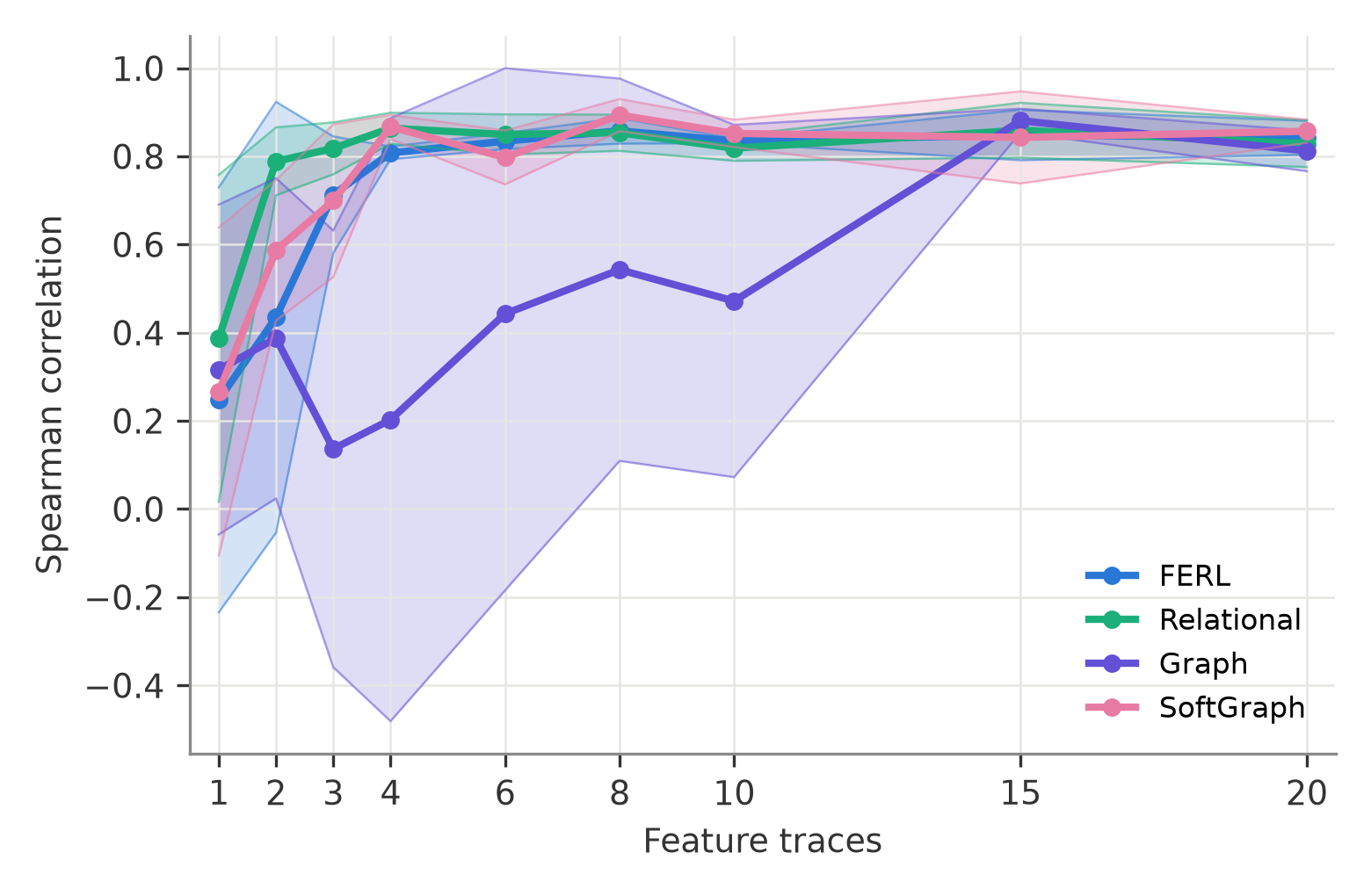}
    }
    \subfigure[Environment MSE]{
        \includegraphics[width=0.3\textwidth]{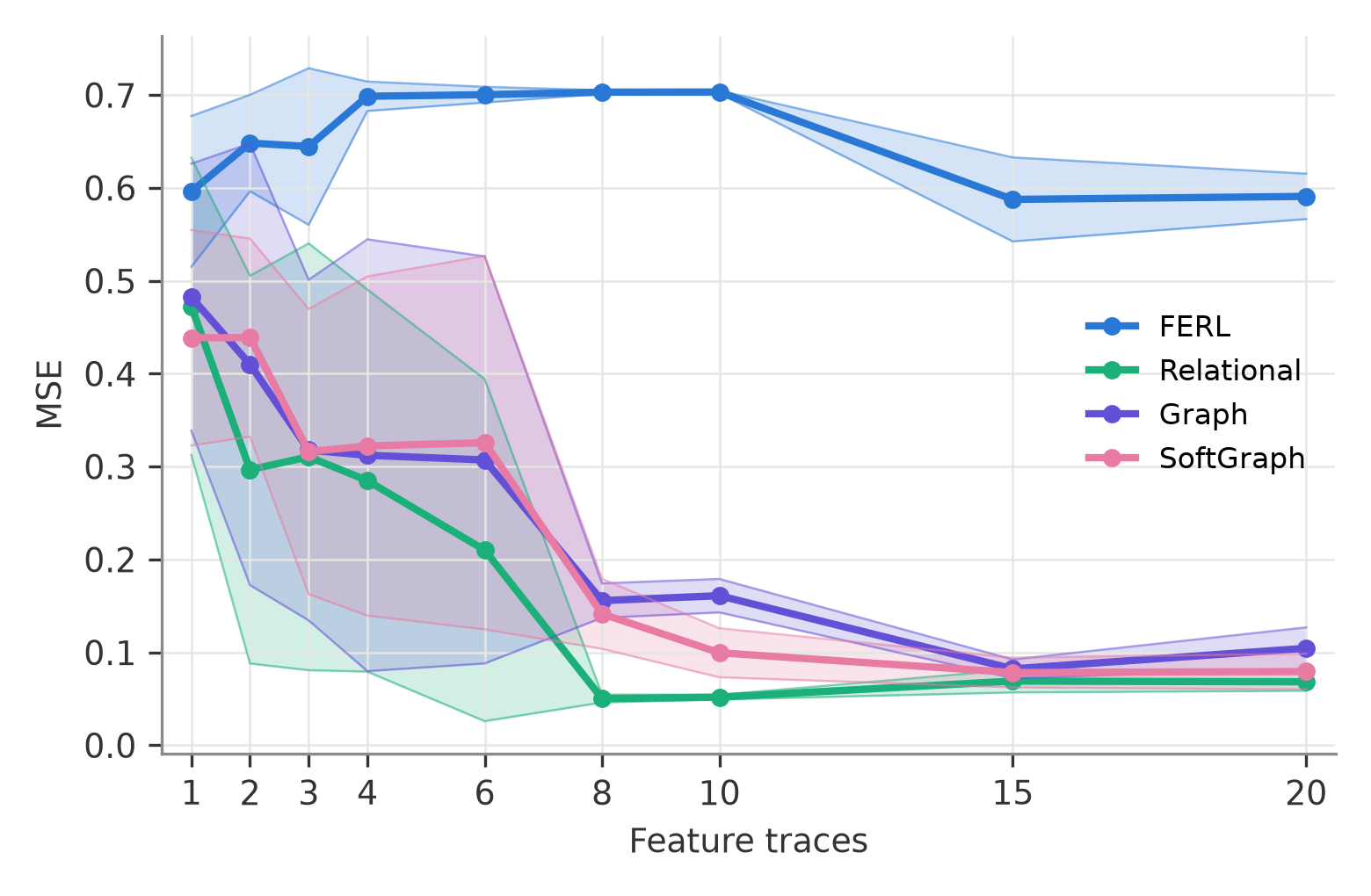}
    }
    \subfigure[Environment PR-AUC]{
        \includegraphics[width=0.3\textwidth]{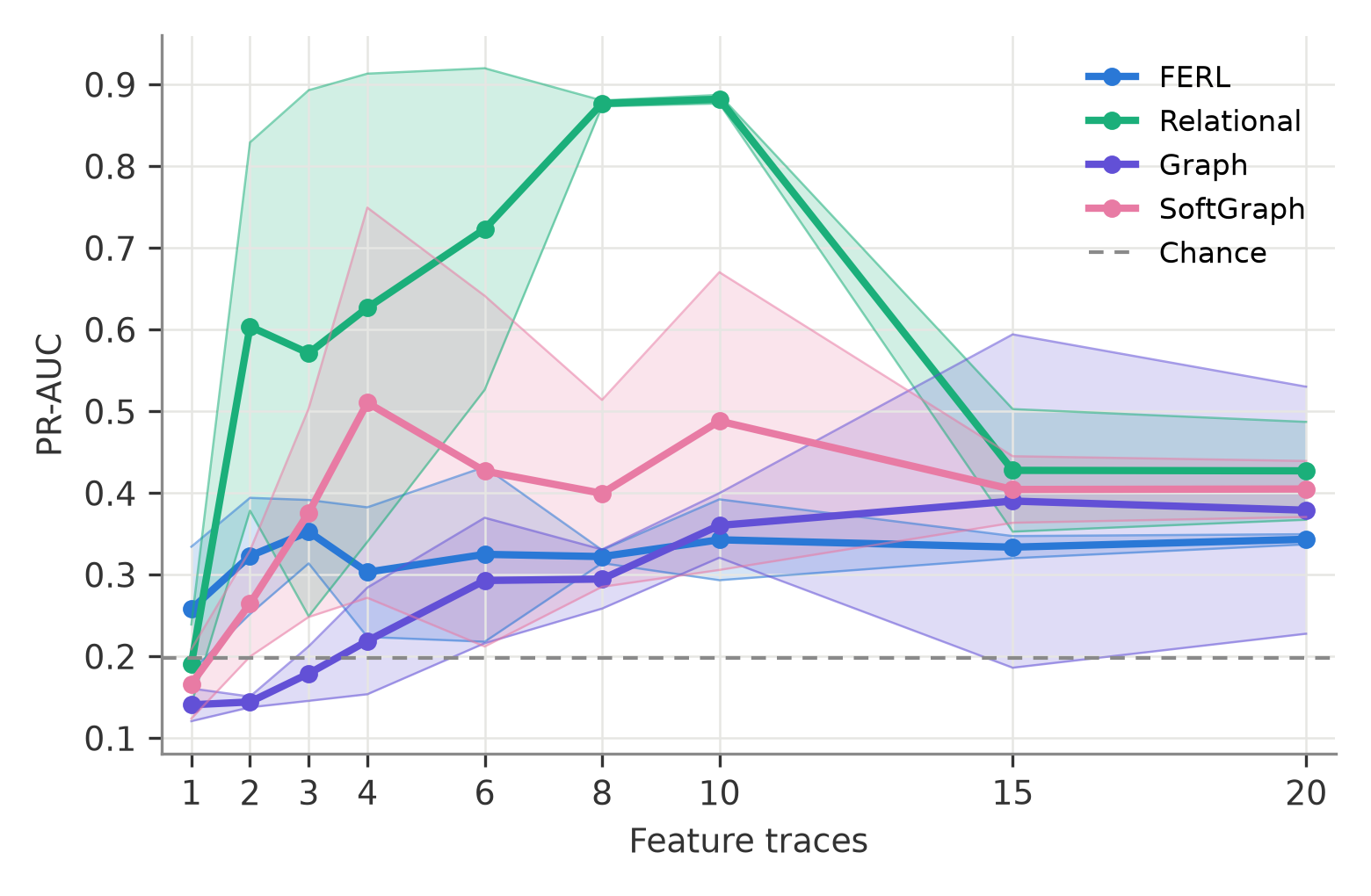}
    }
    \subfigure[Environment Spearman]{
        \includegraphics[width=0.3\textwidth]{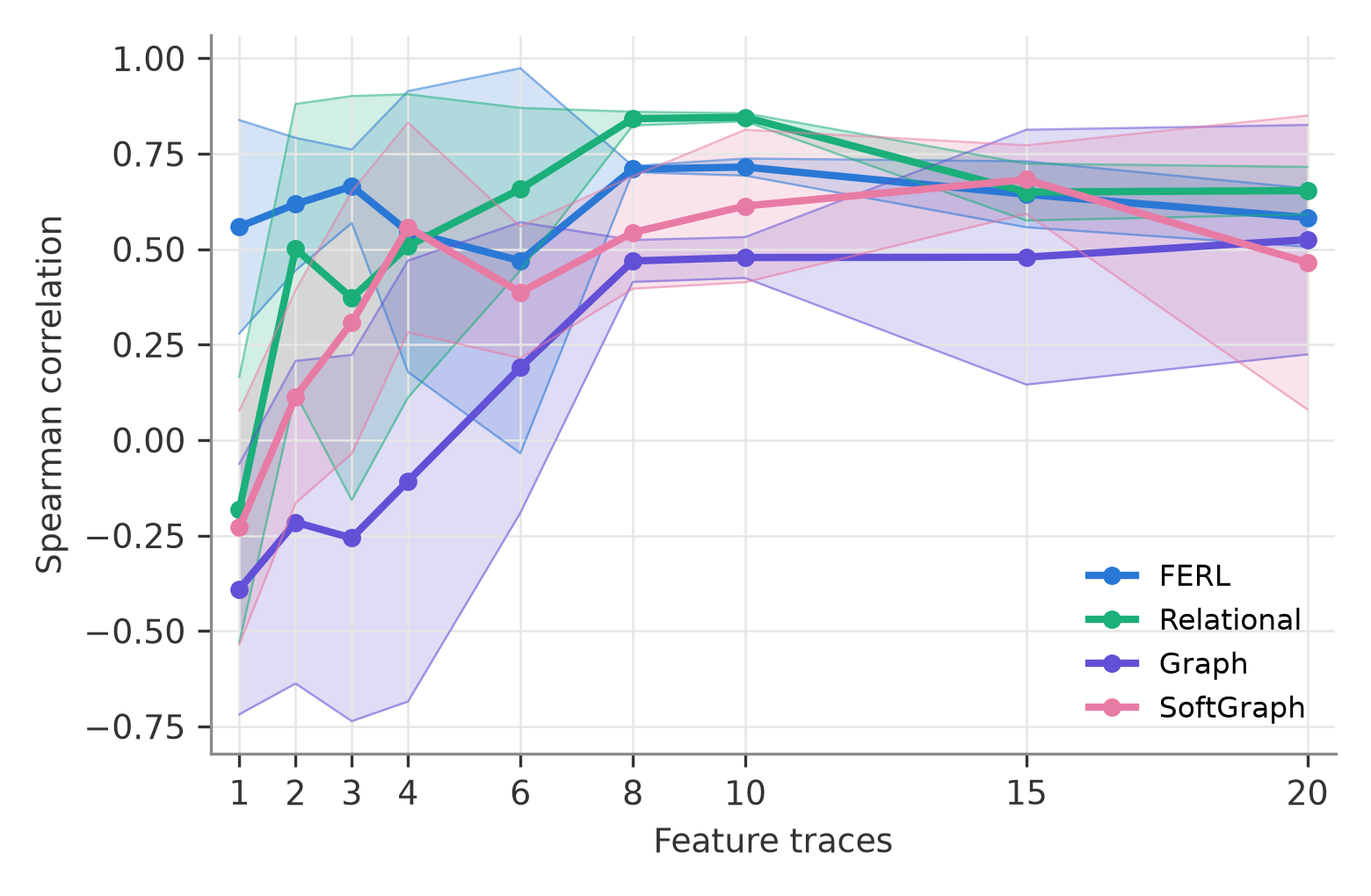}
    }
    \caption{MSE, PR-AUC, and Spearman correlation for the 'avoid laptop' scenario with robot (top row) and environment (bottom row) feature traces.}
    \label{fig:metrics-laptop}
    \vspace{-12pt}
\end{figure}

\begin{figure}[]
    \centering
    \subfigure[Ground Truth]{
        \includegraphics[height=0.18\textwidth]{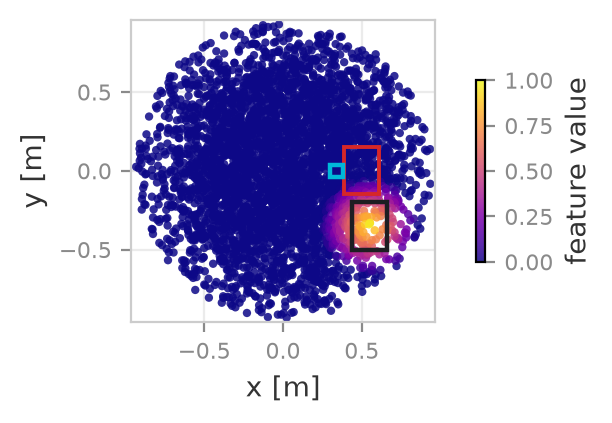}
    }
    \subfigure[Relational Robot]{
        \includegraphics[height=0.18\textwidth]{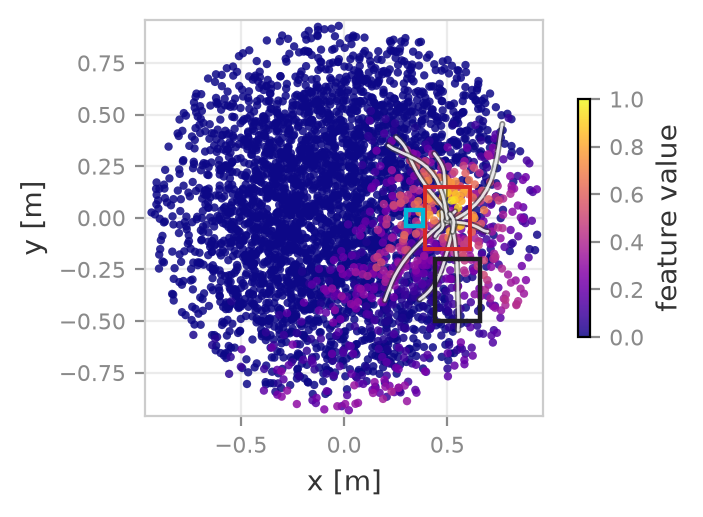}
    }
    \subfigure[Relational Env]{
        \includegraphics[height=0.18\textwidth]{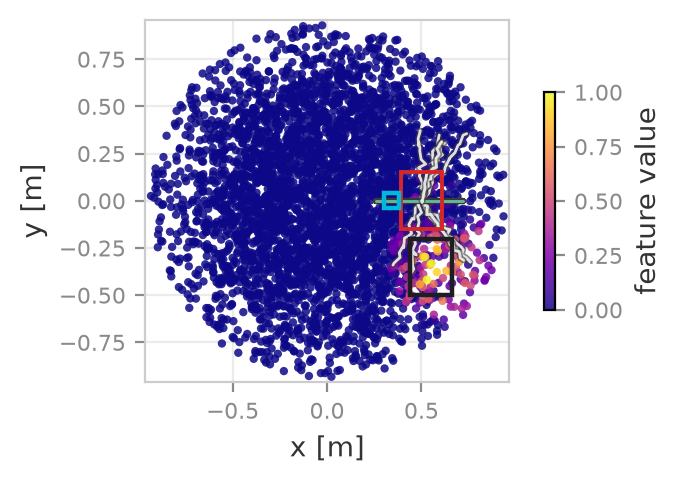}
    }
    \caption{Ground truth and Relational model learned features for the ambiguous 'avoid laptop' scenario. Higher feature values indicate lower reward/higher cost. The laptop and cup's initial positions are shown in red and teal respectively. The laptop's test-time outline is shown in black. Feature traces are shown in white. For environment traces, the robot path is shown in teal.}
    \label{fig:ambiguity}
    \vspace{-12pt}
\end{figure}

\begin{figure}
    \centering
    \subfigure[Robot MSE]{
        \includegraphics[width=0.3\textwidth]{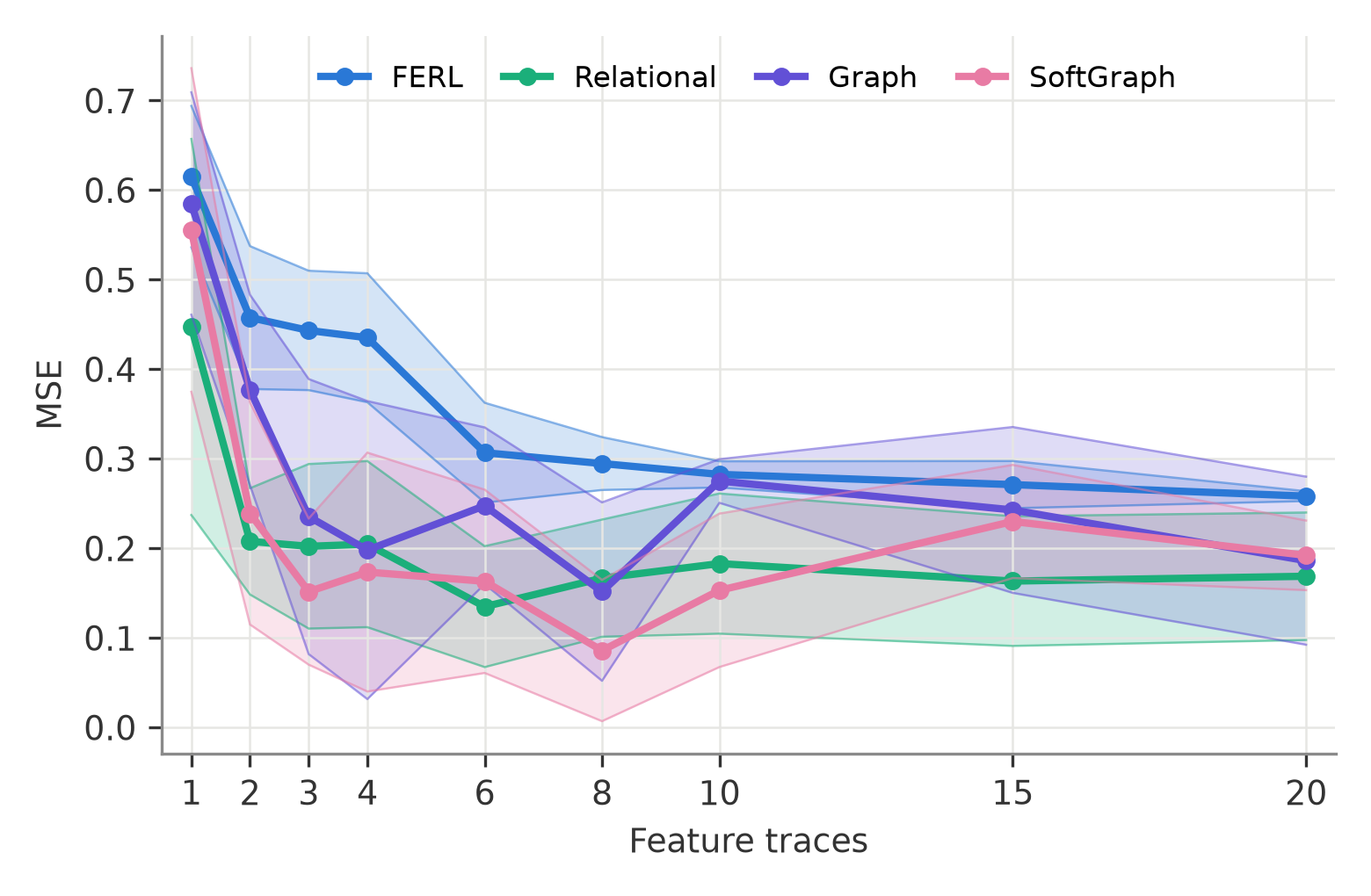}
    }
    \subfigure[Robot PR-AUC]{
        \includegraphics[width=0.3\textwidth]{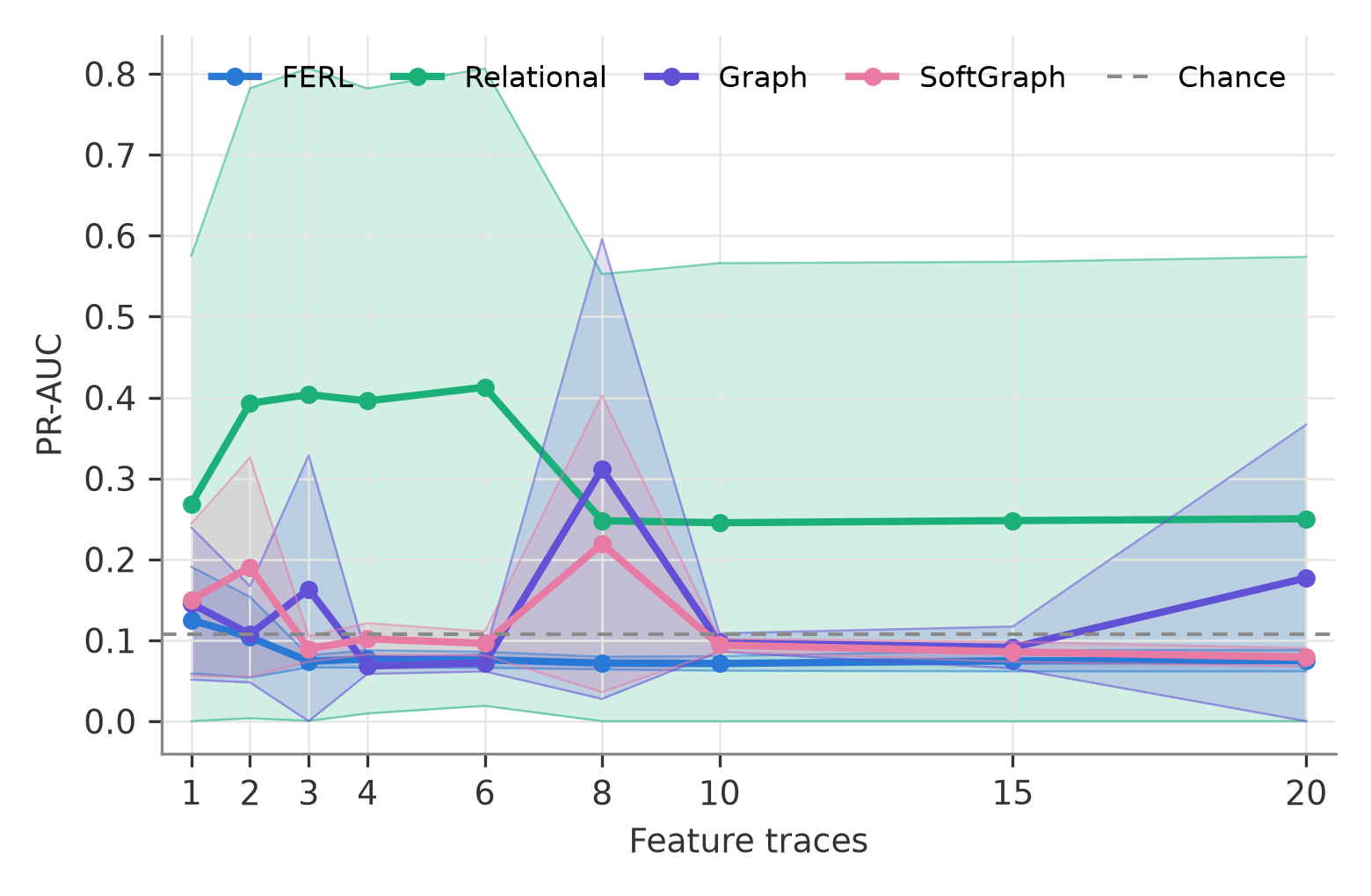}
    }
    \subfigure[Robot Spearman]{
        \includegraphics[width=0.3\textwidth]{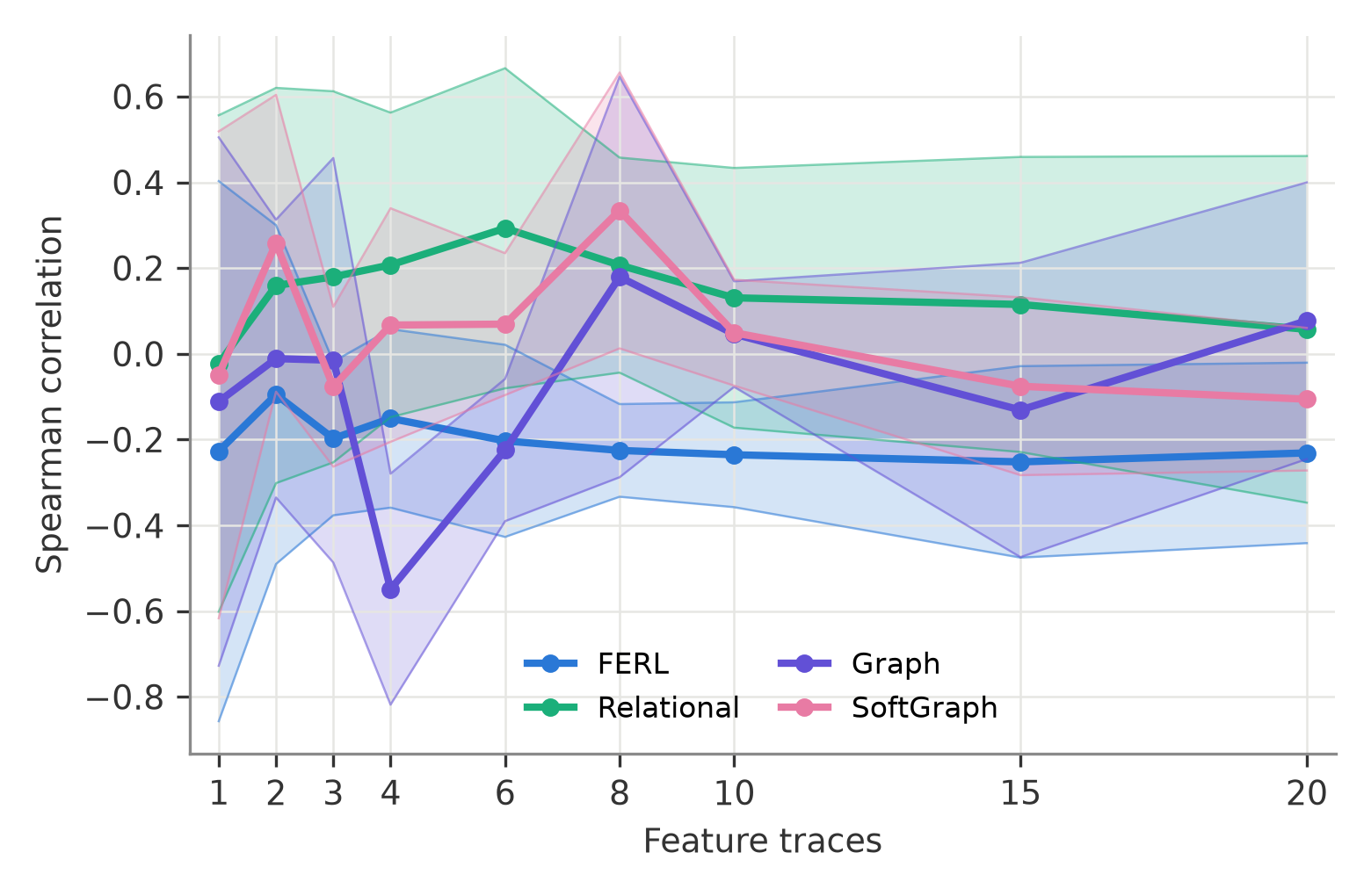}
    }
    \subfigure[Environment MSE]{
        \includegraphics[width=0.3\textwidth]{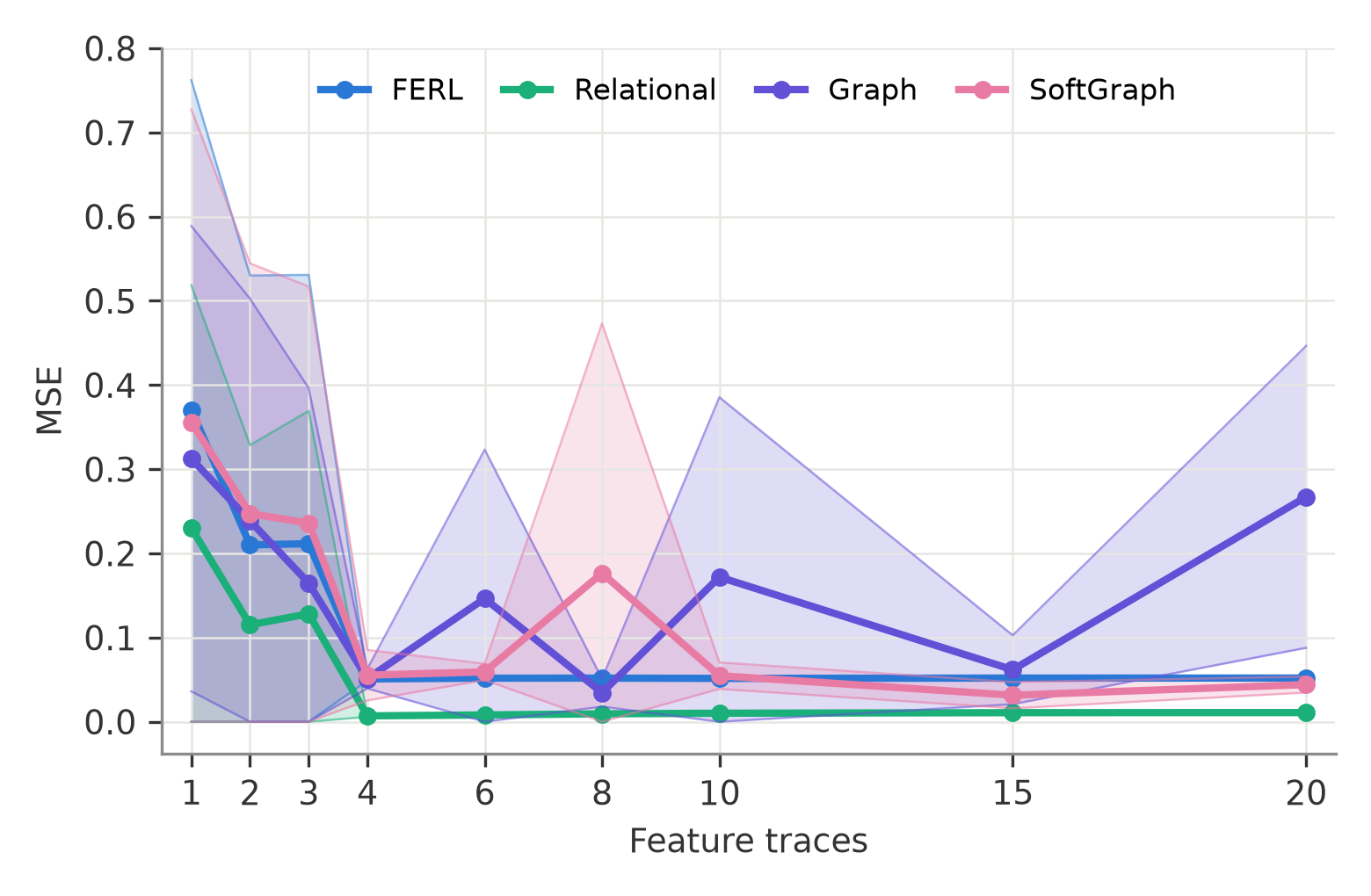}
    }
    \subfigure[Environment PR-AUC]{
        \includegraphics[width=0.3\textwidth]{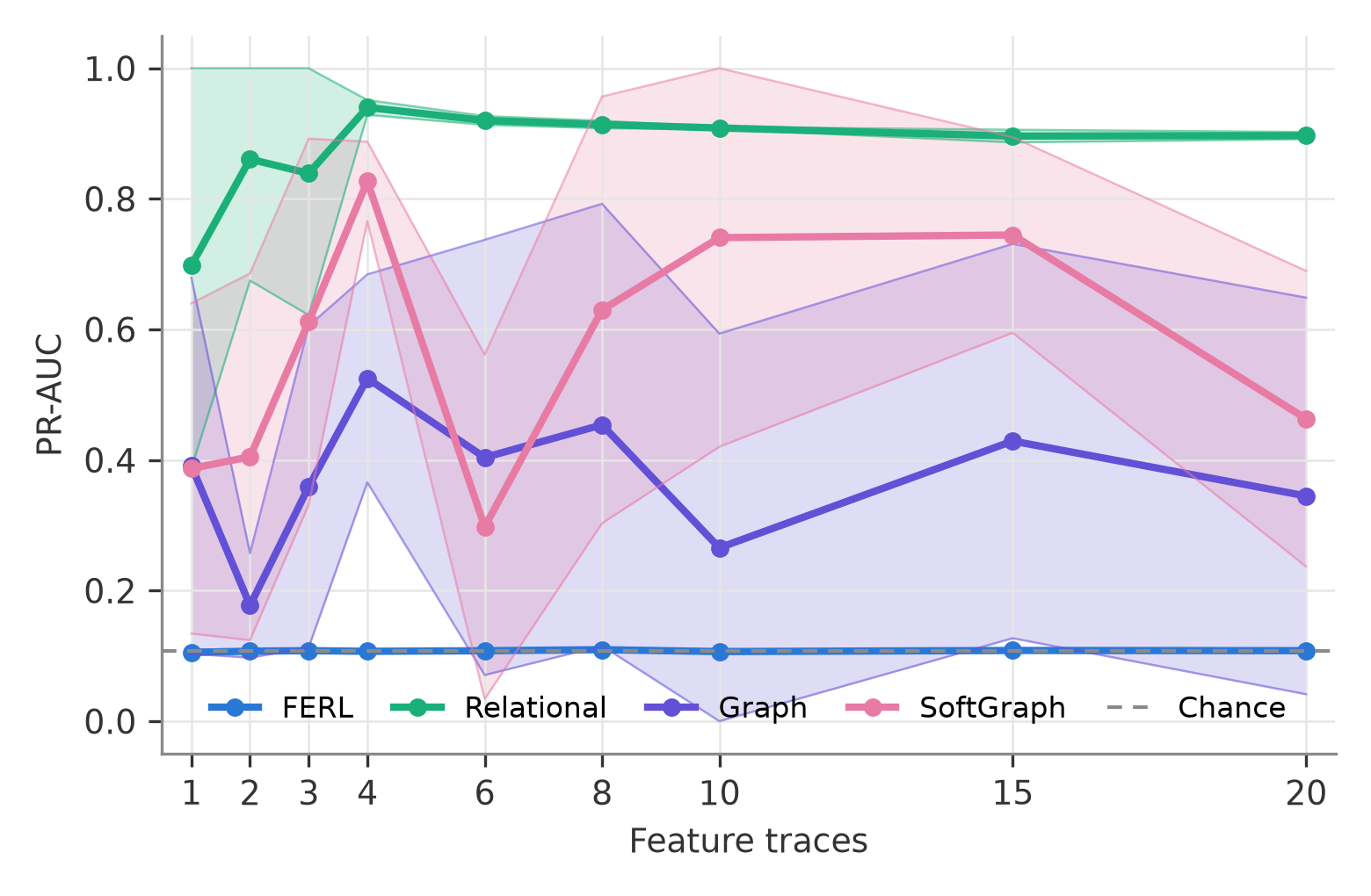}
    }
    \subfigure[Environment Spearman]{
        \includegraphics[width=0.3\textwidth]{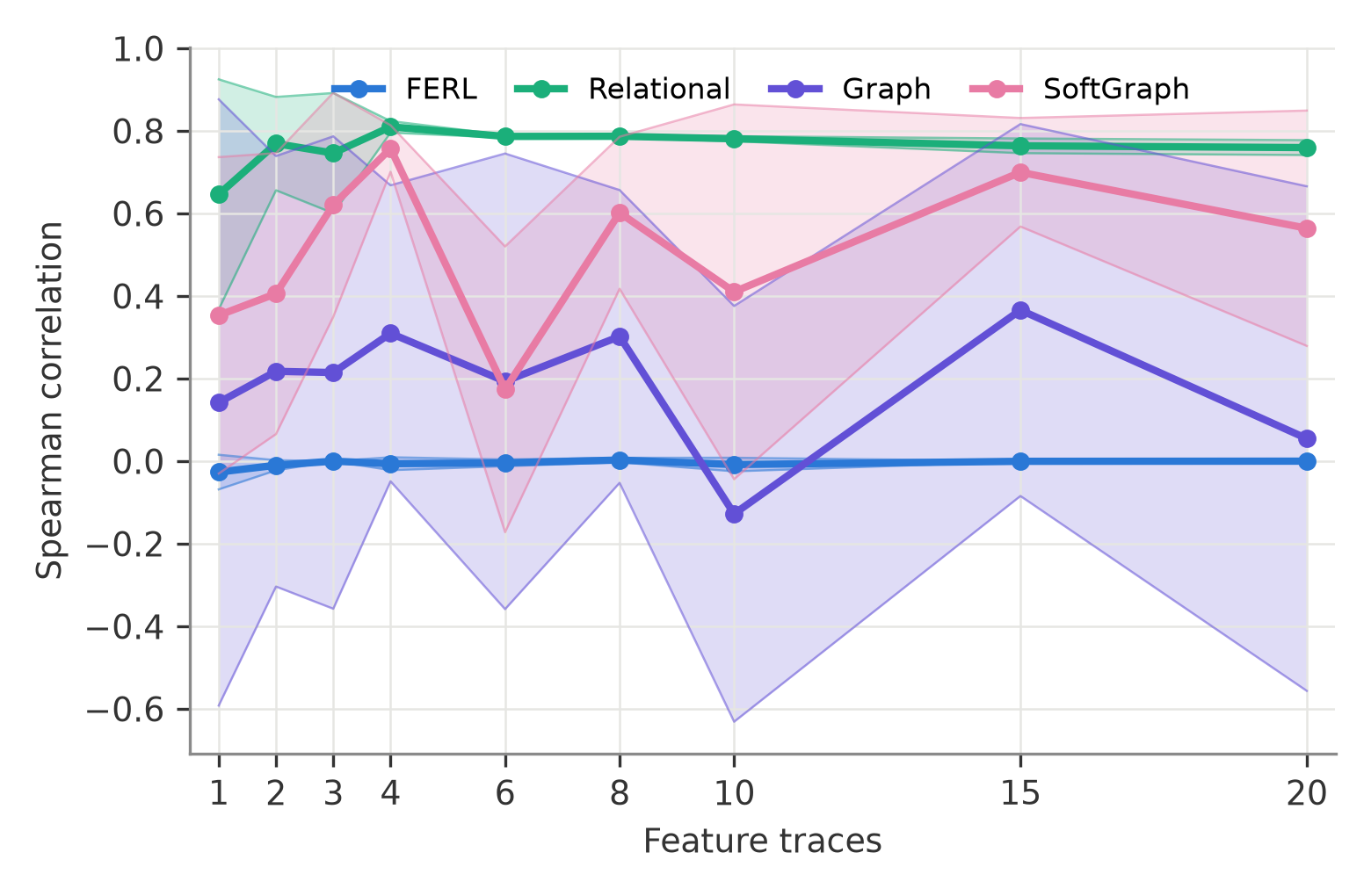}
    }
    \caption{MSE, PR-AUC, and Spearman correlation for the ambiguous 'avoid laptop' scenario with robot (top row) and environment (bottom row) feature traces.}
    \label{fig:metrics-ambiguous}
    \vspace{-12pt}
\end{figure}

\begin{figure}
    \centering
    \subfigure[Ground Truth (Open Cup)]{
        \includegraphics[height=0.20\textwidth]{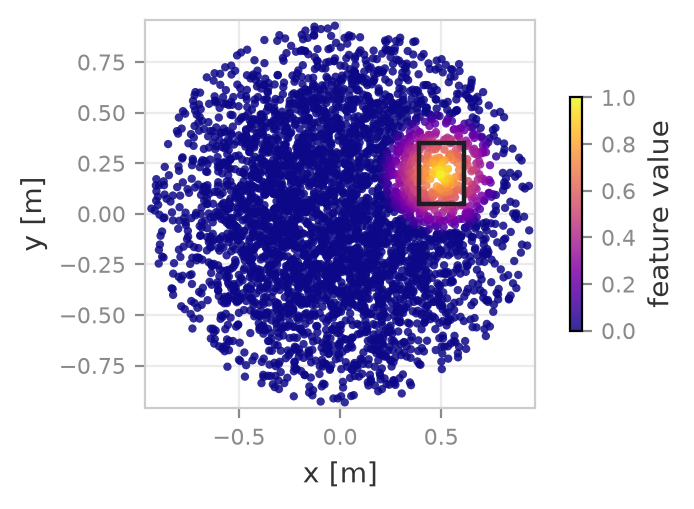}
    }
    \subfigure[SoftGraph (Open Cup)]{
        \includegraphics[height=0.20\textwidth]{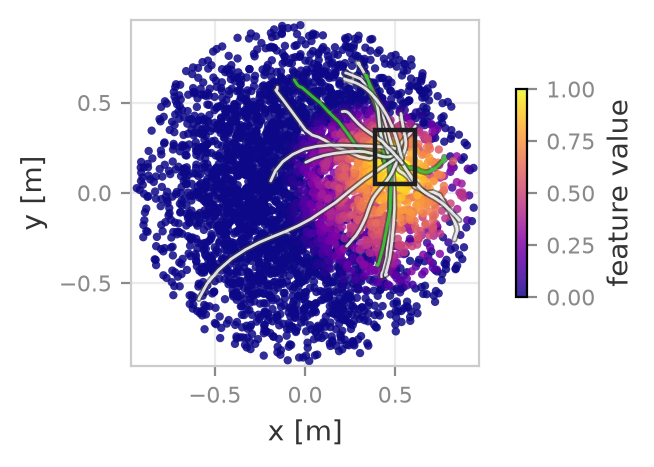}
    }
    \subfigure[SoftGraph (Closed Cup)]{
        \includegraphics[height=0.20\textwidth]{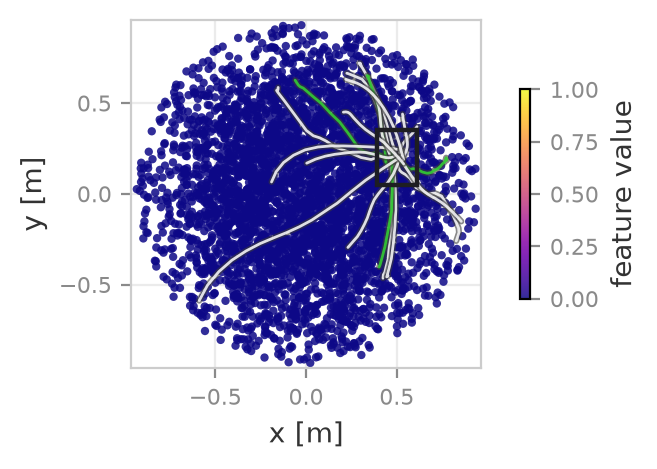}
    }
    \caption{Ground truth and learned rewards for the feature 'avoid laptop when carrying an open cup'. Rewards are shown when evaluated for the robot carrying an open and a closed cup. Robot feature traces are overlaid on the reward. 'No preference' feature traces for the closed cup are shown in green. The ground truth for the closed cup is uniformly zero.}
    \label{fig:open_vs_closed}
    \vspace{-12pt}
\end{figure}

\begin{figure}
    \centering
    \subfigure[MSE (Closed Cup)]{
        \includegraphics[width=0.23\textwidth]{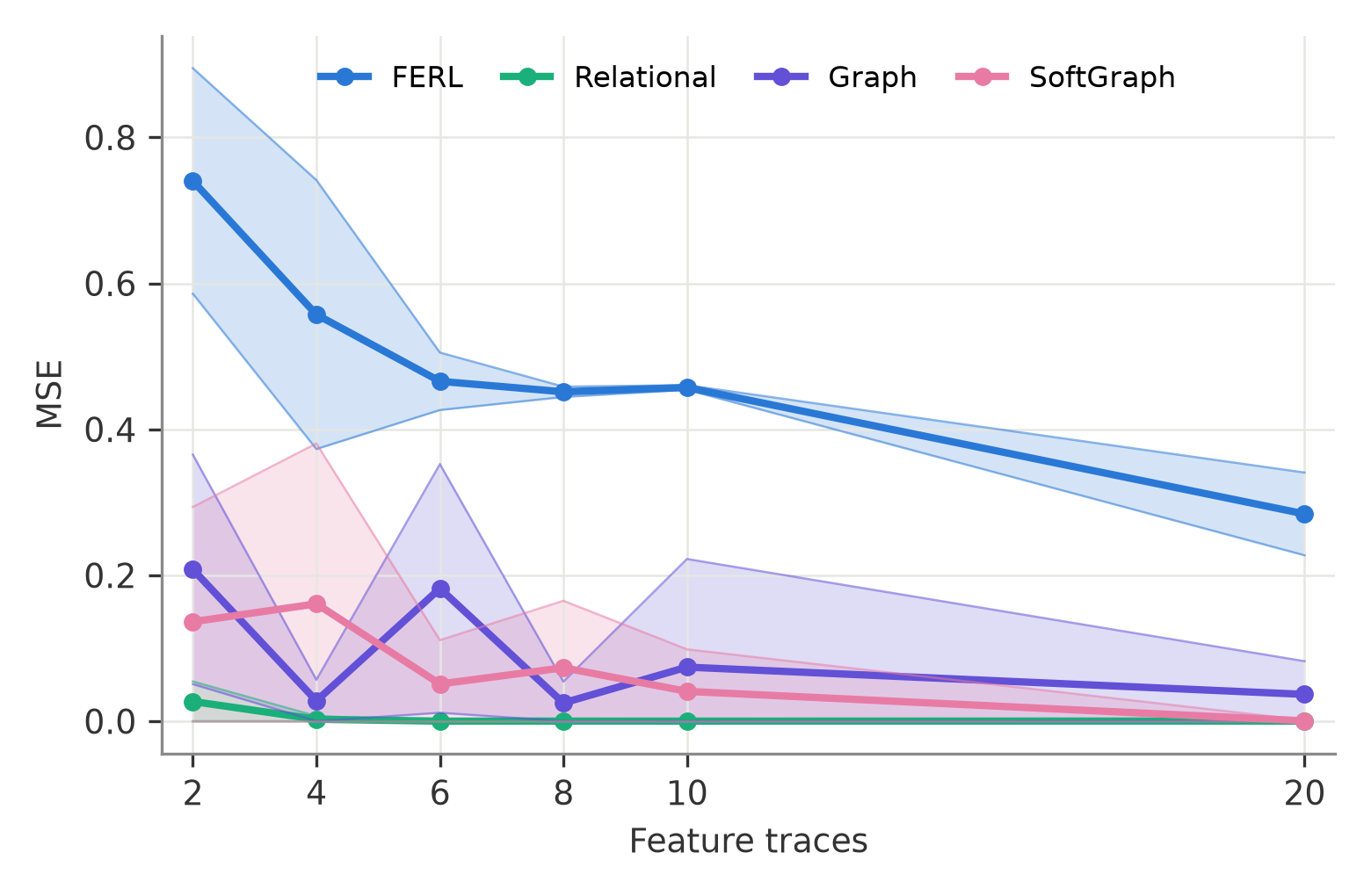}
    }
    \subfigure[MSE (Open Cup)]{
        \includegraphics[width=0.23\textwidth]{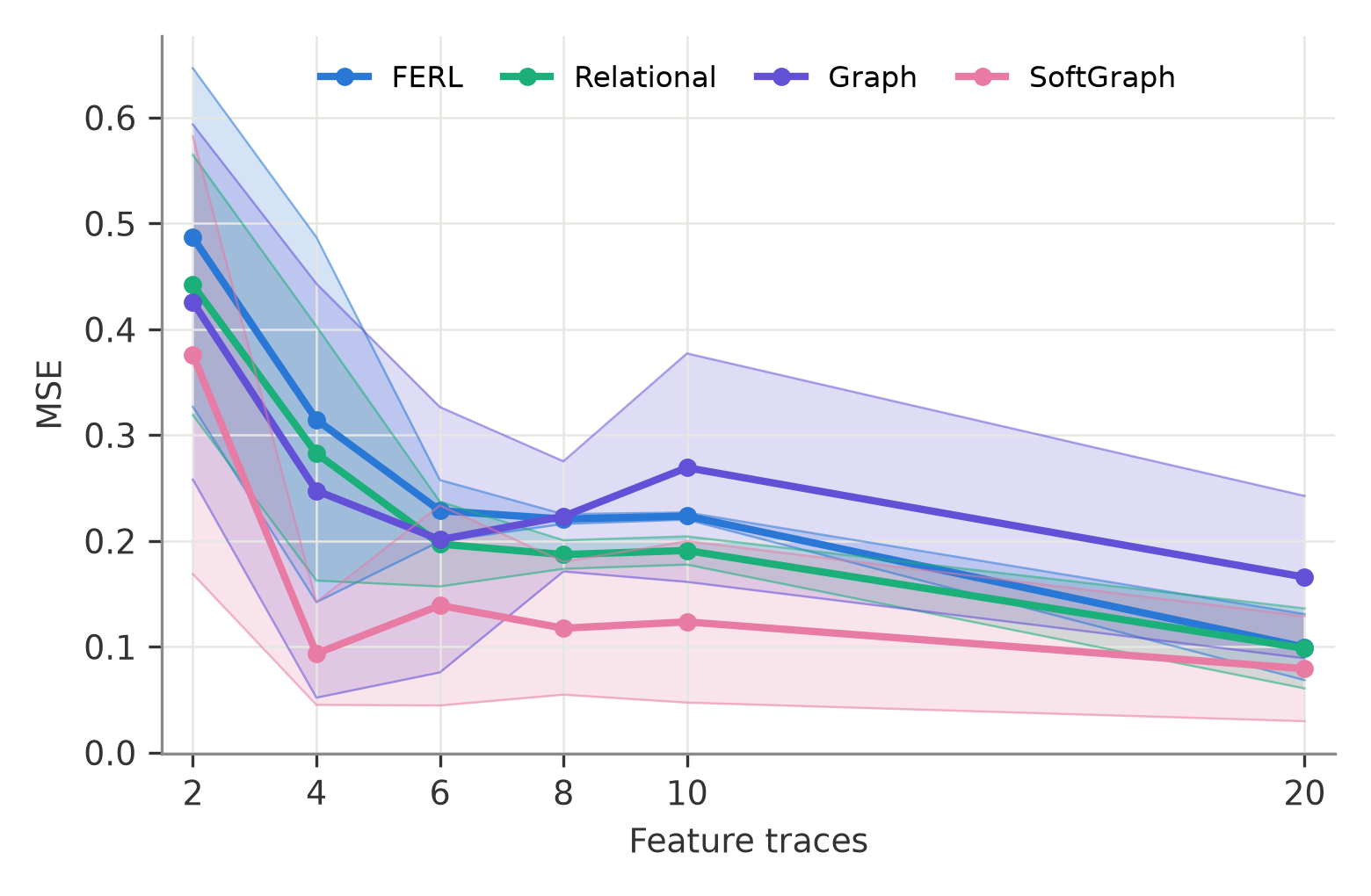}
    }
    \subfigure[PR-AUC (Open Cup)]{
        \includegraphics[width=0.23\textwidth]{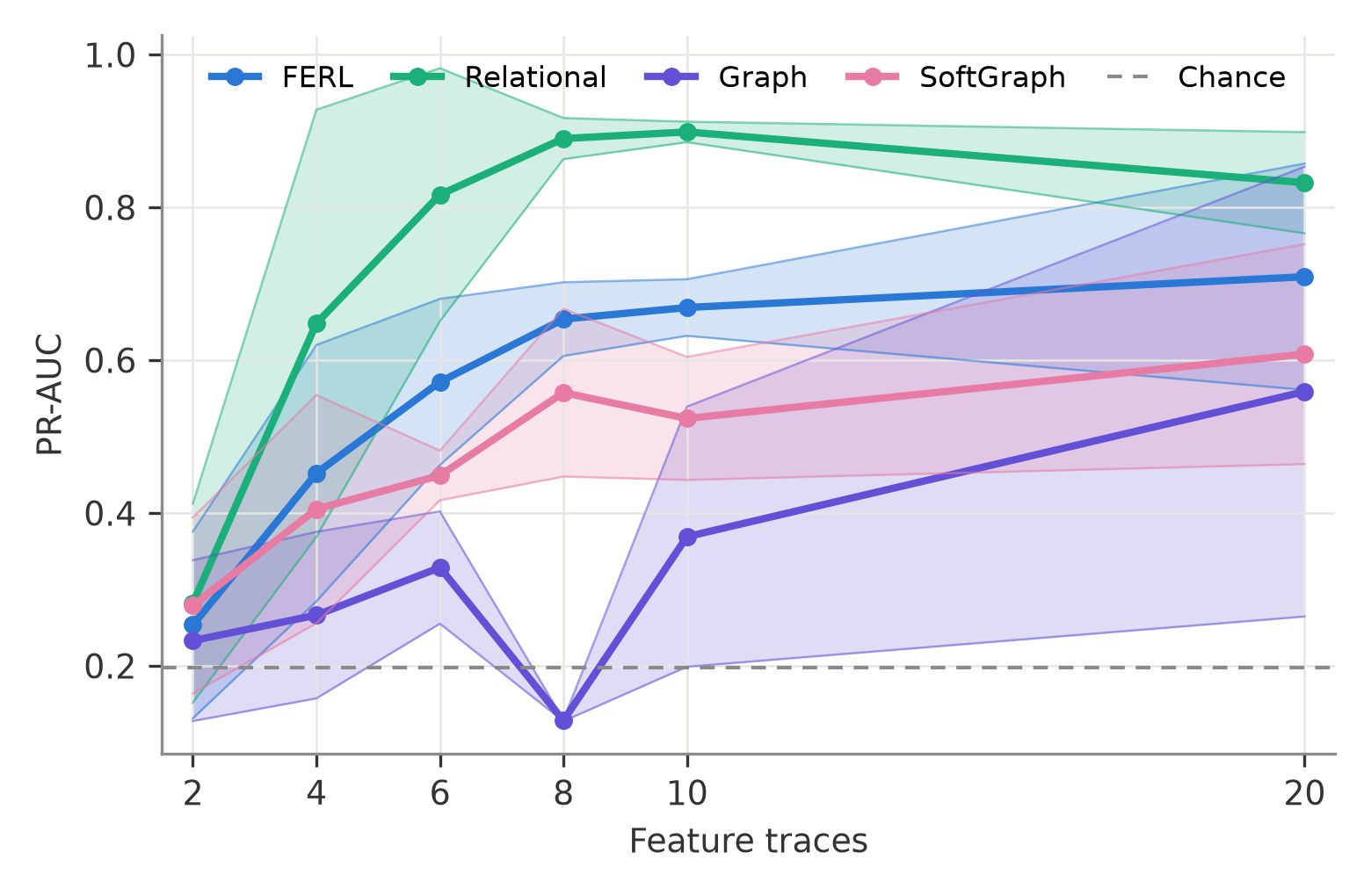}
    }
    \subfigure[Spearman (Open Cup)]{
        \includegraphics[width=0.23\textwidth]{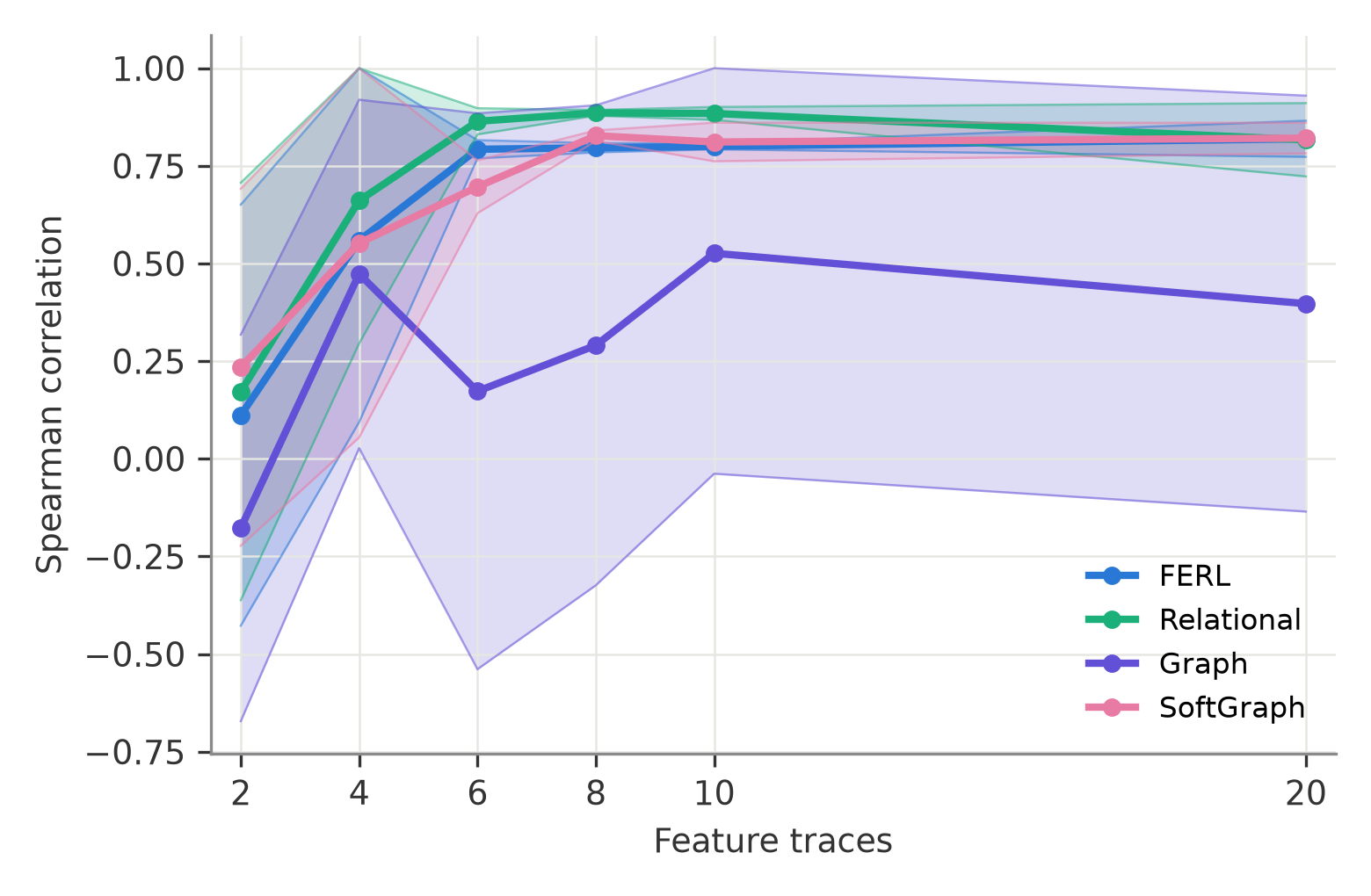}
    }
    \caption{MSE for both open and closed cup conditions, and PR-AUC and Spearman correlation for the open cup condition.}
    \label{fig:metrics-open-vs-closed}
    \vspace{-12pt}
\end{figure}


\section{Experimental Results}
\label{sec:experiments}

\subsection{Experimental Setup}
We consider three scenarios and compare our proposed model architectures against FERL~\cite{bobu2021ferl}. One scenario compares using robot and environmental feature traces to indicate that the robot should not move over a laptop. The next goes a step further to consider an ambiguous environment configuration in which there is a cup next to the laptop that does not need to be avoided. The third explores a conditional feature where the laptop only needs to be avoided when the robot is carrying an open cup and not a closed one.
We consider scaling numbers of feature traces, which were generated by the authors. For each number of feature traces, five seeds are used. We report mean squared error (MSE) between the ground truth feature, area under the precision-recall curve (PR-AUC) where a feature value of 0.4 or higher is considered a high-cost area, and the Spearman correlation.

\subsection{Learning Object Proximity Features}
Figure~\ref{fig:laptop} shows 
the ground truth and an example of FERL's learned features using both robot and environment feature traces over the arm's reachable space, projected to 2D. Figure~\ref{fig:metrics-laptop} shows the MSE, PR-AUC, and Spearman correlation.
From robot feature traces, all models converge to learning the feature, though the Graph model requires more data to achieve this.
Considering environmental traces, FERL's input does not contain enough information about the scene to learn the feature values; however, the PR-AUC and Spearman correlation support that it is able to learn small regions of higher feature values directly over the laptop.
The Graph and SoftGraph models learn a much wider area of high feature expression than the ground truth. The Relational model is the only one that learns a consistent feature model between robot and environmental edits.

\subsection{Disambiguating Spurious Correlations}
Figure~\ref{fig:ambiguity} shows 
the ground truth and an example of the Relational model's learned features for the ambiguous scenario. Figure~\ref{fig:metrics-ambiguous} shows the MSE, PR-AUC, and Spearman correlation.
From the robot corrections, no model was able to learn that the feature related to the laptop itself and all learned high feature expression near the initial position of the laptop and the cup. Using environment traces, which moved the laptop away, the Relational model is able to learn a feature similar to the ground truth. The graph models again, similar to the previous scenario, learned wider areas of high feature expression than the ground truth.
FERL was unable to associate the feature with the laptop.

\subsection{Attribute-Specific Features}
Figure~\ref{fig:open_vs_closed} shows 
the ground truth for the open cup condition of avoiding the laptop when carrying an open cup and not a closed cup. The ground truth for the closed condition is uniform over the space. Figure~\ref{fig:metrics-open-vs-closed} shows the MSE for both conditions and PR-AUC and Spearman correlation for the open condition.
For this, we modified each model to support 'no preference' traces where the loss function encourages each state in the trace to have an equal feature value.
We provided robot feature traces while carrying an open cup and an equal number of 'no preference' traces while carrying a closed cup.
FERL does not take object attributes and learns the same feature for both conditions. The Relational, Graph, and SoftGraph models learn similar features for the open and closed conditions, with the Graph and SoftGraph models recovering close to zero error on the closed condition. This may indicate that graph models are able to better learn feature dependence on object attributes.


\section{Conclusion}
\label{sec:conclusion}
We introduce environment edits as a corrective signal for reward learning and present FERRET, which learns features from both robot and environment traces.
We show that FERL is unable to learn features that require information from environment traces to disambiguate spurious correlations.
We explore two different architectures for implementing FERRET, a relational model and two variants of a graph model. Our experiments show that the Relational model's hand-built input features make it consistent across trace types, while the graph models learn from the scene geometry itself and are better able to capture attribute dependent preferences but are prone to over-wide feature expression.
Considering future work, we plan to explore replacing hand-specified attributes with learned object embeddings, using boundary-of-acceptability edits to recover feature weights, and conduct a user study to explore how people naturally choose between correcting the robot and correcting the scene.
	

\clearpage
\acknowledgments{
This work was supported in part by the IBM-Illinois Discovery Accelerator Institute, the Center
for Networked Intelligent Components and Environments (C-NICE), and by the Illinois-Insper Partnership at the University of Illinois.
}


\bibliography{refs}  

\end{document}